\documentclass[letterpaper, 10 pt, conference]{ieeeconf}  

\IEEEoverridecommandlockouts                              

\usepackage{hyperref}
\hypersetup{
    colorlinks,
    citecolor=black,
    filecolor=black,
    linkcolor=black,
    urlcolor=black
}

\usepackage{times}
\usepackage{amsmath}
\usepackage{comment}
\usepackage{gensymb}
\usepackage{calc}
\usepackage[T1]{fontenc}
\usepackage{amssymb}
\usepackage{amsmath}
\usepackage{amstext}
\usepackage{graphicx,color}
\usepackage{booktabs,tabularx}
\usepackage{multirow}
\usepackage{url}
\usepackage{tabularx}
\graphicspath{{images/}}
\usepackage{caption}
\usepackage{subcaption}
\usepackage{tabularx} 
\usepackage{wrapfig}
\usepackage{caption}
\usepackage[normalem]{ulem}

\makeatletter
\let\NAT@parse\undefined
\renewcommand{\@IEEEsectpunct}{ }
\makeatother
\usepackage[numbers, sort]{natbib}

\newcolumntype{Y}{>{\centering\arraybackslash}X}

\newcommand{\ignore}[1]{}

\def\secref#1{Sec.~\ref{#1}}
\def\figref#1{Fig.~\ref{#1}}
\def\tabref#1{Tab.~\ref{#1}}
\def\eqref#1{Eq.~(\ref{#1})}

\DeclareMathOperator*{\dist}{dist}

\DeclareMathOperator*{\minimize}{\textrm{minimize}}

\newcommand{\se}[1]{\mathit{SE}(#1)}
\newcommand{\so}[1]{\mathit{SO}(#1)}
\newcommand{\realNumbers}{\mathbb{R}}

\newcommand{\transpose}{^\top}
\newcommand{\object}{o}

\newcommand{\pointCloud}{\mathbf{P}}
\newcommand{\objectPointCloud}[1]{\pointCloud_{#1}}
\newcommand{\point}{\mathbf{p}}
\newcommand{\objectPoint}[1]{\point_{#1}}
\newcommand{\centroid}{\mathbf{c}}
\newcommand{\objectCentroid}[1]{\centroid_{#1}}

\newcommand{\transform}{\mathbf{T}}
\newcommand{\pose}[1]{\transform_{#1}}

\newcommand{\relativePose}[2]{^{#2}\transform_{#1}}
\newcommand{\relativePoseOptimal}[2]{\relativePose{#1}{#2}^*}
\newcommand{\rotMat}{\mathbf{R}}

\newcommand{\rotMatRel}[2]{^{#2}\rotMat_{#1}}
\newcommand{\transVec}{\mathbf{t}}

\newcommand{\transVecRel}[2]{^{#2}\transVec_{#1}}
\newcommand{\scene}{\mathbf{s}}
\newcommand{\testScene}{\scene^*}
\newcommand{\sceneNthDemo}[1]{\scene^{(#1)}}
\newcommand{\demos}{\mathcal{D}}
\newcommand{\nnScenes}{\demos_{\mathit{NN}}}
\newcommand{\nnRatioGood}{\epsilon_{\mathit{NN}}}
\newcommand{\nnRatioThresh}{\epsilon_{*}}
\newcommand{\numSceneNN}{Q}
\newcommand{\relationFeatureNthDemo}[1]{\relationFeatureVector^{(#1)}}
\newcommand{\newRelationDemos}{\demos'}
\newcommand{\onlineMetricDemos}{\demos^*}

\newcommand{\stateSet}{\mathcal{S}}
\newcommand{\feasibleSet}{\stateSet_{\mathrm{feas}}}
\newcommand{\featureFunction}{f}
\newcommand{\relationFeatureVector}{\mathbf{r}}
\newcommand{\relationFeatureVectorDim}{K}
\newcommand{\testSceneFeature}{\relationFeatureVector^*}
\newcommand{\setOfRelationFeautres}{\mathcal{R}}
\newcommand{\setOfNewRelationFeautres}{\setOfRelationFeautres'}
\newcommand{\newRelationFeautreExample}{\relationFeatureVector'}
\newcommand{\setOfFriends}[1]{\setOfRelationFeautres^+_{#1}}
\newcommand{\setOfImposters}[1]{\setOfRelationFeautres^-_{#1}}

\newcommand{\gravity}{\mathbf{g}}
\newcommand{\normal}{\mathbf{n}}

\newcommand{\histogram}{\mathbf{h}}
\newcommand{\angA}{\theta}
\newcommand{\angB}{\varphi}
\newcommand{\histAngA}{\histogram_{\angA}}
\newcommand{\histAngB}{\histogram_{\angB}}
\newcommand{\histDist}{\histogram_{d}}
\newcommand{\simMatrix}{\mathbf{Y}}
\newcommand{\simValue}{y}
\newcommand{\simValueRowCol}[2]{\simValue_{#1,#2}}

\newcommand{\metricMatrixDecomposed}{\mathbf{L}}

\newcommand{\chiSq}{\chi^2}
\newcommand{\lmnn}{LMNN}
\newcommand{\lmnnMargin}{\zeta}
\newcommand{\gblmnn}{GB-LMNN}
\newcommand{\chiSqlmnn}{$\chiSq$\textrm{-LMNN}}
\newcommand{\distTwoScenes}[2]{\dist(\relationFeatureVector_{#1}, \relationFeatureVector_{#2})}

\newcommand{\distParams}{\phi}
\newcommand{\priorDistParams}{\distParams_0}
\newcommand{\onlineDistParams}{\distParams_*}
\DeclareMathOperator*{\distPhi}{dist_{\distParams}}
\DeclareMathOperator*{\priorMetric}{dist_{\priorDistParams}}
\DeclareMathOperator*{\onlineMetric}{dist_{\onlineDistParams}}

\newcommand{\pushPullConst}{\lambda}

\newcommand{\lossFunction}{\mathcal{L}}

\usepackage{flushend}

\newcommand{\papertitle}{Metric Learning for Generalizing Spatial Relations to New Objects}

\title{\LARGE \bf
\papertitle
}

\author{Oier Mees \and Nichola Abdo  \and Mladen Mazuran \and Wolfram Burgard
  \thanks{
    All authors are with the Department of Computer Science, University of Freiburg, Germany. 
    \{meeso, abdon, mazuran, burgard\}@informatik.uni-freiburg.de  }
}
\begin{document}
\maketitle
\thispagestyle{empty}
\pagestyle{empty}

\begin{abstract}
Human-centered environments are rich with a wide variety of spatial
relations between everyday objects. For autonomous robots to operate
effectively in such environments, they should be able to reason about these
relations and generalize them to objects with different shapes and sizes. For
example, having learned to place a toy inside a basket, a robot should be able to
generalize this concept using a spoon and a cup. This
requires a robot to have the flexibility to learn arbitrary relations
in a lifelong manner, making it challenging for an 
expert to pre-program it with sufficient knowledge to do so beforehand. In this
paper, we address the problem of learning spatial relations by introducing a novel method from the perspective
of distance metric learning. Our approach enables a robot to reason about the
similarity between pairwise spatial relations, thereby enabling it to use its previous
knowledge when presented with a new relation to imitate. We show how this makes 
it possible to learn arbitrary spatial relations from non-expert users using a 
small number of examples and in an interactive manner. Our extensive evaluation
with real-world data demonstrates the effectiveness of our method in reasoning
about a continuous spectrum of spatial relations and generalizing them to new
objects.
\end{abstract}

\IEEEpeerreviewmaketitle

\section{Introduction}
\label{sec:introduction}


Understanding spatial relations is a crucial faculty of autonomous robots
operating in human-centered environments. We expect future service robots to
undertake a variety of everyday tasks such as setting the table, tidying up, or 
assembling furniture. In this context, a robot should be able to reason about the best way
to reproduce a spatial relation between two objects, e.g., by placing an item
inside a drawer, or aligning two boxes side by side.

However, our everyday environments typically include a rich spectrum of
potential spatial relations. For example, each user may have different
preferences with respect to object arrangements, which requires robots to be 
flexible enough to handle arbitrary relations they have not encountered
before. Similarly, robots should be able to generalize relations they have
learned and achieve them using new objects of different shapes or sizes. 
For these reasons, it is highly impractical to expect an expert to pre-program 
a robot with the knowledge it needs to handle all potential situations in the
real world, e.g., in the form of symbols. 
Instead, we aim for a lifelong learning approach that enables non-expert users
to teach new spatial relations to robots in an intuitive manner.

One way to do this is to provide a robot with several examples using 
different objects in order to learn a model for a new relation, e.g., ``inside.'' 
On the one hand, this may require generating large amounts of data to learn the
new relation, which is impractical in setups in which a
robot learns from a non-expert teacher. On the other hand, this requires
learning a new model for each relation individually, making it hard for the
robot to reuse its knowledge from previous relations.

\begin{figure}[t]
  \centering
\includegraphics[width=0.44\textwidth]{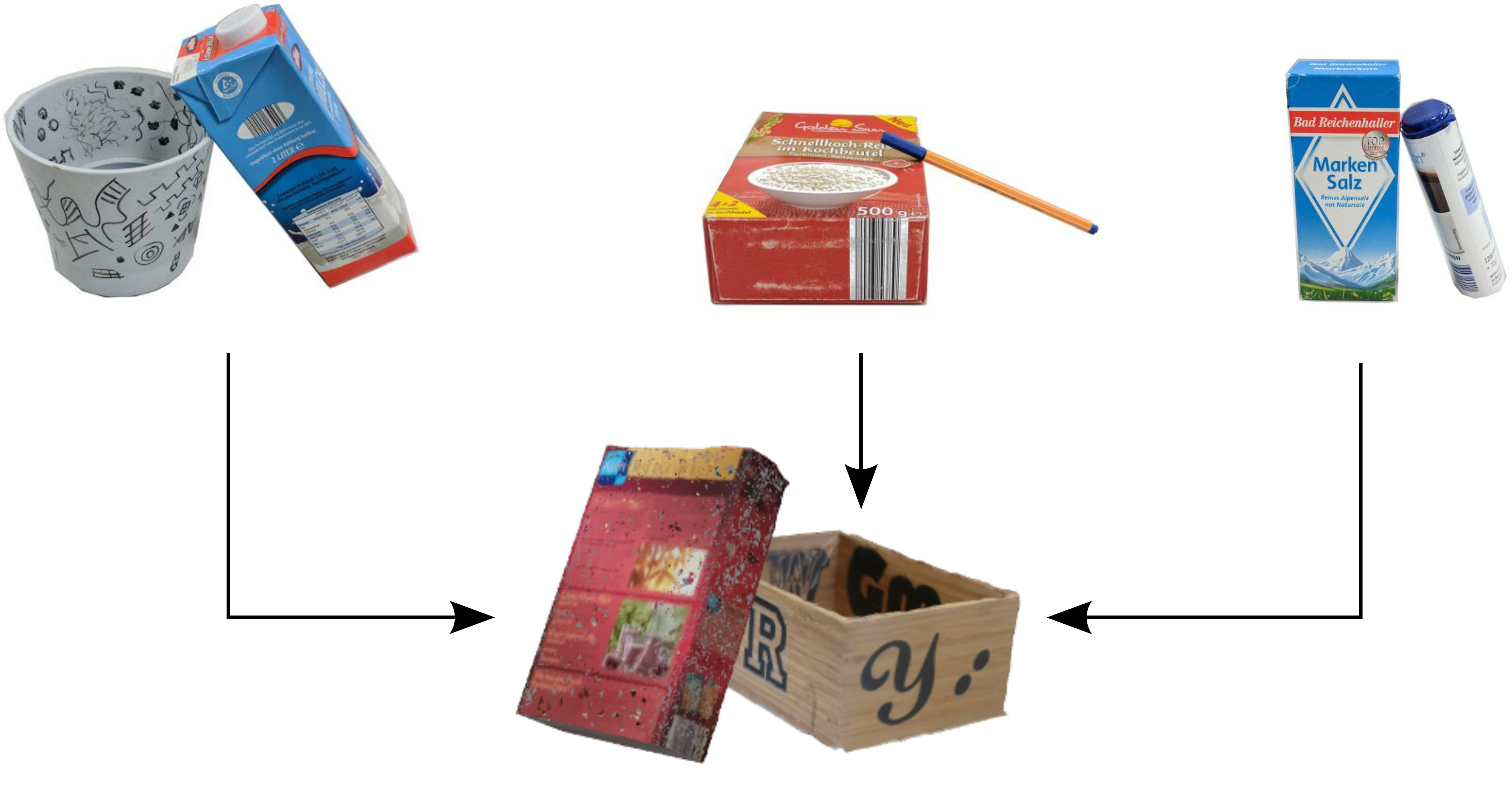}
 
 \caption{\label{fig:coverGirl}We present a novel method based on distance
         metric learning to reason about the similarity between 
         pairwise spatial relations. Our approach uses demonstrations of a
 relation given by non-expert teachers (top row) in order to generalize this
 relation to new objects (bottom row).}
\end{figure}
In this paper, we address this problem from the perspective of distance metric
learning and focus on learning relations between pairs of objects. 
We present a novel method that allows the robot to reason about how
similar two relations are to each other. By doing so, we formulate the problem
of reproducing a relation using two new objects as one of minimizing the
distance between the reproduced relation and the teacher demonstrations. More
importantly, our approach enables the robot to use a few teacher
demonstrations as queries for retrieving \emph{similar} relations it has seen before,
thereby leveraging prior knowledge to bootstrap imitating the new relation. 
Therefore, rather than learning a finite set of individual relation models, 
our method enables reasoning on a continuous spectrum of relations.

Concretely, we make the following contributions: \emph{i)} we present a novel 
approach from the perspective of distance metric learning to address the problem of 
learning pairwise spatial relations and generalizing them to new objects\footnote{The Freiburg Spatial Relations Dataset and a demo video of our approach running on the PR-2 robot are available at \url{http://spatialrelations.cs.uni-freiburg.de}}, 
\emph{ii)} we introduce a novel descriptor that encodes pairwise spatial
relations based only on the object
geometries and their relative pose, \emph{iii)} we demonstrate how our method 
enables bootstrapping the
learning of a new relation by relating it to similar, previously-learned
relations, \emph{iv)} we present an interactive learning method that enables non-expert users to teach arbitrary
spatial relations from a small number of examples, and \emph{v)} we
present an extensive evaluation of our method based on real-world data we
gathered from different user demonstrations.






%

\section{Related Work}
\label{sec:relatedWork}



In the context of robotics, previous work has focused on leveraging 
predefined relations in the form of symbolic predicates for solving tasks, 
as in the case of combined task and motion planning or in
the context of relational reinforcement learning~\cite{kaelbling2013integrated,
hofer2016coupled, paxton2016want, beetz2010cram}. Rather than relying on grounding existing
symbols, other works have addressed learning symbols  and
effects of actions to abstract continuous states for the purpose of high-level
planning~\cite{abdo2013learning, konidaris2014constructing, pasula2007learning,
jetchev2013learning, ahmadzadeh2015learning}. As opposed to these works, we reason about 
the similarity between relations by
learning the distance between scenes, allowing us to compute scenes
that generalize a relation to new objects.

Related to this is the work by~\citeauthor{rosman2011learning}, which proposes
constructing a contact point graph to classify spatial relations \cite{rosman2011learning}. 
Similarly, \citeauthor{fichtl2014learning} 
train random forest classifiers for relations based on histograms that encode
the relative position of surface patches~\cite{fichtl2014learning}. 
\citeauthor{guadarrama2013grounding} learn models of pre-defined
prepositions by training a multi-class logistic regression model using data
gathered from crowdsourcing~\cite{guadarrama2013grounding}. 
As opposed to those works, we propose learning a distance metric that captures the
similarities between different relations without specifying explicit classes.


Moreover, related to our work is the interactive approach
by~\citeauthor{kulick2013active} for learning relational symbols from a
teacher~\cite{kulick2013active}. They use Gaussian Process classifiers to model
symbols and therefore enable a robot to query the teacher with examples to
increase its confidence in the learned models. Similarly, our method enables a robot to generalize a relation by
interacting with a teacher. However, we do this from the perspective of metric
learning, allowing the robot to re-use previous demonstrations of other
relations.




Similar to our work,~\citeauthor{zampogiannis2015learning} model spatial
relations based on the geometries of objects given their point cloud
models~\cite{zampogiannis2015learning}. However, they define a variety
of common relations and focus on addressing the problem of extracting the
semantics of manipulation actions through temporal analysis of
spatial relations between objects.
Other methods have also relied on the
geometries of objects and scenes to reason about preferred object
placements~\cite{jiang2012learning} or likely places to find an object~\cite{aydemir2012exploiting}.
Moreover, \citeauthor{kroemer2014predicting} used 3D object models to extract contact point 
distributions for predicting interactions between objects~\cite{kroemer2014predicting}.

Finally, our approach leverages distance metric learning for reasoning about
the similarity between relations. Metric learning is a popular paradigm in the
machine learning and computer vision communities. Learned metrics have been
applied to address face recognition \cite{guillaumin2009you}, image
classification \cite{hoi2008semi} and image segmentation
\cite{xiang2008learning}. In the context of robotics, metric learning has been
used to address problems related to object instance or place
recognition~\cite{lai2011sparse, shahid16rss_ws}.
%

\section{Notation and Problem Formulation}
\label{sec:problemFormulation}
In this section, we formalize the problem we address in this paper.
\subsection{Object Representation}
\label{sec:stateRepresentation}
We consider the problem of learning spatial relations between pairs of objects.
We denote an object by $\object$. In this work, we assume to have no semantic
knowledge about objects such as their type, e.g., box. Instead, we aim to learn
relations based on object geometries and assume to have a 3D model of each
object $\object_k$ in the form of a point
cloud $\objectPointCloud{k}$. We consider only points on the surface of the objects. We model the state using the 3D poses of
objects in $\se{3}$ and express the pose $\pose{k}$ of $\object_k$ relative to the world
frame as a homogeneous transform consisting of a translation vector $\transVec_k\in\realNumbers^3$ 
and a rotation matrix $\rotMat_k\in\so{3}$. 
We denote the pose of $\object_l$ relative to $\object_k$ by
$\relativePose{l}{k}$. Additionally, we assume the world frame to
be specified such that the $-z$-axis aligns with the gravity vector $\gravity$.

\subsection{Pairwise Relations}
\label{sec:pairwiseRelations}
We consider learning pairwise spatial relations between objects, i.e., we
consider scenes with two objects only. In this work, we do not address the
perception problem and rely on existing techniques to segment the scene and compute 
the object poses based on their point clouds. Accordingly, we express a 
scene with $\object_k$ and $\object_l$ as a tuple $\scene:=\langle\objectPointCloud{k},\objectPointCloud{l},\,\relativePose{l}{k}\rangle$.
     In this work, we assume that one of the objects ($\object_k$) is
     labeled as the \emph{reference object}, and therefore express the scene using
            the pose $\relativePose{l}{k}$ of $\object_l$ relative to
            $\object_k$.
     Given a scene $\scene$, we aim to express the spatial relation
            between the two objects in it.
     For this, we rely on a feature function (descriptor)
            $\featureFunction$ to express the relation as a $\relationFeatureVectorDim$-dimensional
            feature vector $\relationFeatureVector$, i.e.,
            $\featureFunction(\scene) =
            \relationFeatureVector\in\realNumbers^\relationFeatureVectorDim$.
     Moreover, our goal is to enable the robot to reason about the
            how \emph{similar} two scenes are with respect to the pairwise
            relations in them.
     We capture the similarity between $\scene_1$ and $\scene_2$ using a distance
            function $\dist$ that computes the distance 
            $\distTwoScenes{1}{2}\ge0$ 
            between the two scenes
            with respect to their
            feature vectors $\relationFeatureVector_1$ 
            and $\relationFeatureVector_2$. In this work we do not consider relations involving more than two objects, as they can be defined by combining pairwise relations. Finally, we do not explicitly treat object symmetries.

\subsection{The Problem}
\label{sec:theProblem}


The problem we address is threefold.
%
\subsubsection{Representing relations:}
\label{sec:descriptorProblem}
First, we seek a descriptor $\featureFunction$ that enables us to capture the
underlying spatial relation in a scene based only on the geometries (point
clouds) of the objects, their relative poses, and the direction of gravity
$\gravity$.

\subsubsection{Learning the distance between relations:}
\label{sec:distanceProblem}
Given $\featureFunction$, we aim to learn a distance metric $\dist$ for
computing the distance between two scenes. For this, we rely on training data $\demos=\{\sceneNthDemo{1}, \dots, \sceneNthDemo{N}\}$ consisting of $N$ demonstrated scenes. Additionally, we assume to
have a symmetric similarity matrix $\simMatrix$ of size $N\times N$ with unit
diagonal values. The value $\simValueRowCol{i}{j}$ in the $i$-th row and $j$-th
column of $\simMatrix$ captures the degree of similarity between scenes $\sceneNthDemo{i}$
            and $\sceneNthDemo{j}$ in $\demos$. In this work, we consider
            binary similarities
            $\simValue\in \{0,1\}$, such that 0 
            represents dissimilar relations and 1 means that the relations in
            both scenes are identical.
     Note that we do not assume $\simMatrix$ to be
            completely specified, i.e., some entries may be missing. Therefore,
            we aim for a method that can learn with partially-labeled data with
            respect to scene similarities. 
            Given $\demos$ and $\simMatrix$, our goal is to learn a \emph{distance
            metric} $\dist$ that captures the distance between scenes
            $\scene_1$ and $\scene_2$ based on their features
            $\relationFeatureVector_1 = \featureFunction(\scene_1)$ and
            $\relationFeatureVector_2 = \featureFunction(\scene_2)$. We
            learn this metric such that $\distTwoScenes{1}{2}$ is ``small''
            if $\scene_1$ and $\scene_2$ represent similar relations, and
            ``large'' if they represent dissimilar relations.

\subsubsection{Generalizing a relation to new objects:}
\label{sec:imitationProblem}
Given $\featureFunction$ and a distance metric $\dist$, our goal is to learn a new, arbitrary relation from a teacher. We assume the teacher
provides a small set of demonstrations $\newRelationDemos=\{\sceneNthDemo{1},\dots,
\sceneNthDemo{N'}\}$ of the new relation, where $1\leq N' \ll N$. Given two 
new objects $\object_k$ and $\object_l$, the robot has to
``imitate'' the demonstrated relation in
$\newRelationDemos$ by computing the pose $\relativePose{l}{k}$ of
$\object_k$ relative to $\object_l$ such that the
resulting scene $\testScene = \langle\objectPointCloud{k},
\objectPointCloud{l}, \,\relativePose{l}{k}\rangle$ is close to the
demonstrations with respect to the corresponding features.

Concretely, we seek the best $\relativePoseOptimal{l}{k}$ to a problem of 
the form:
\begin{equation}
\label{eqn:imitationProblem}
\begin{aligned}
        &\mathrm{minimize}\quad
        \lossFunction(\setOfNewRelationFeautres,\testSceneFeature)\\
            &\mathrm{subject\,to}\quad \relativePose{l}{k}\in\se{3}, 
            \;\testScene\in\feasibleSet,
\end{aligned}
\end{equation}
where
$\setOfNewRelationFeautres=\{\relationFeatureNthDemo{1},\dots,\relationFeatureNthDemo{N'}\}$
is the set of features for the demonstrated scenes $\newRelationDemos$,
$\testSceneFeature = \featureFunction(\testScene)$ is the feature vector of the
test scene $\testScene$, and $\lossFunction$ is a loss function describing the
distance between the demonstrations and the test scene based on $\dist$.
Additionally, $\feasibleSet$ denotes the set of physically feasible scenes. In
this work, we focus on computing the desired pose between the two objects and
do not consider the problem of manipulating the objects to achieve this pose. 
Therefore, we consider $\feasibleSet$ as the set of scenes in which 
$\object_k$ and $\object_l$ are not colliding.

\section{Proposed Feature Representation}
\label{sec:pipas}

\begin{figure}[t]
\centering
\includegraphics[width=0.28\textwidth]{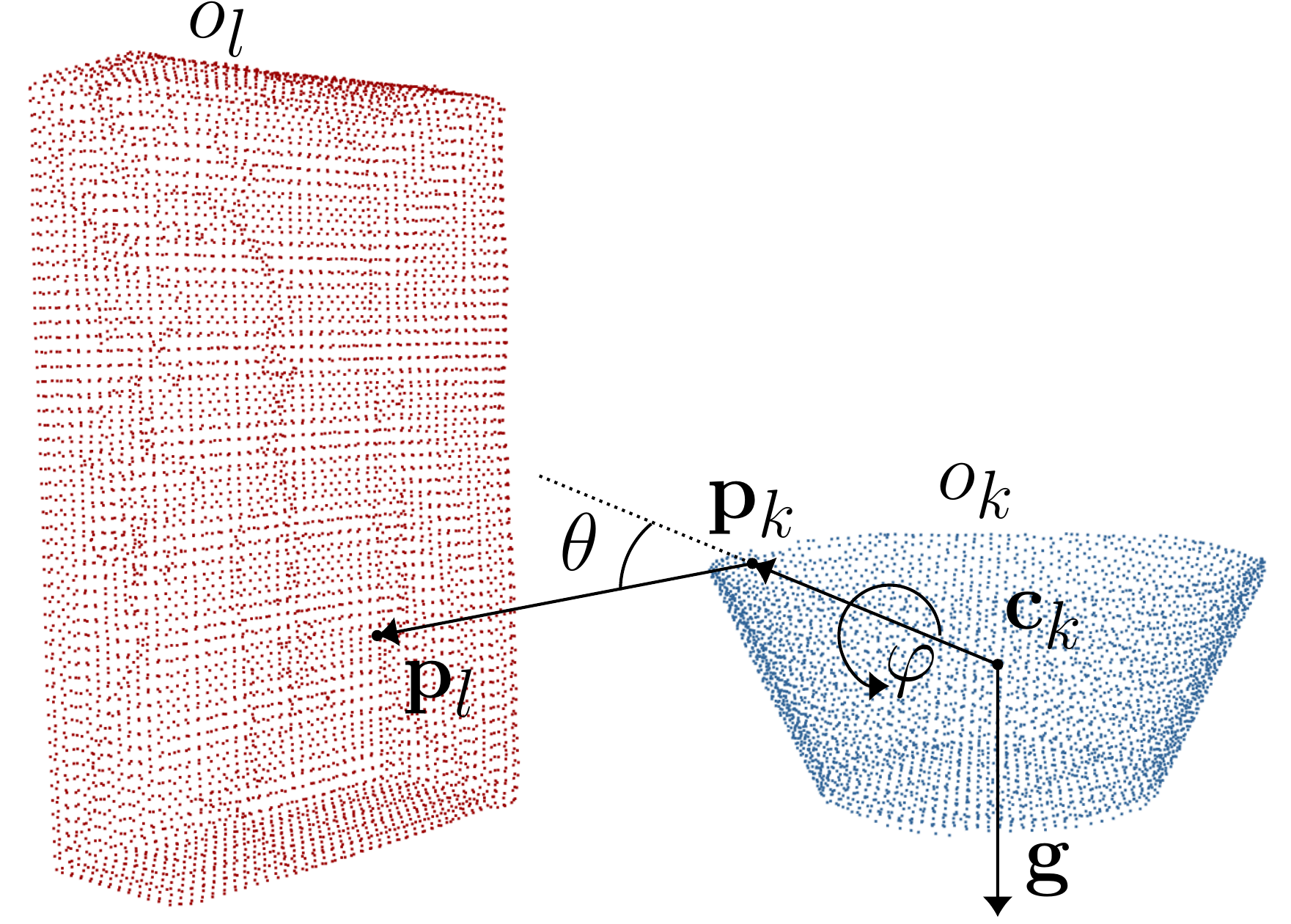}
\caption{\label{fig:pipas} Visualization of the descriptor computation for the spatial relation between objects $\object_k$ and $\object_l$. We consider the gravity vector
        $\gravity$ at the centroid $\mathbf{c}_k$ of the 
reference object $\object_k$. We compute angles $\angA$ and $\angB$ based on
direction vectors involving all surface points $\objectPoint{k}$ and $\objectPoint{l}$ on
$\object_k$ and $\object_l$, respectively.}
\end{figure}

In this section, we present our model for
$\featureFunction$ and propose a descriptor for modeling 
pairwise spatial relations, thereby addressing~\secref{sec:descriptorProblem}. 
We model relations based only on the spatial interaction between their point clouds
$\objectPointCloud{k}$ and $\objectPointCloud{l}$ given the 
direction of the gravity vector $\gravity$. Note that in this work, we do not address the correspondence problem between scenes
        and assume the teacher specifies the reference object.
We rely on the directions of the vectors 
between the surface points of the objects as a signature of the underlying relation 
between them. Defining these directions purely based on a
fixed (world) reference frame is sub-optimal as this results in a descriptor 
that is affected by translations and rotations of
the scene. At the same time, computing a local reference frame
using one of the objects (e.g., using PCA) introduces the challenge of
ensuring consistency and reproducibility of the axes across different scenes.
We address this problem by computing angles between direction vectors between 
points on both objects and the centroid of the reference object
$\object_k$, see~\figref{fig:pipas}. This is analogous to methods 
for computing 
rotationally-invariant descriptors for 2D images such as RIFT~\cite{rift}.  Accordingly, we propose a descriptor that is based on three 
histograms as follows:

\begin{equation}
\label{eqn:pipas}
f(\scene) := [\histAngA\quad \histAngB\quad \histDist]\transpose.
\end{equation}

With the first histogram $\histAngA$, we capture the angular relation 
between the two objects regardless of how the scene is oriented in the global
reference frame. We construct $\histAngA$ as a distribution over the angle $\theta$
between vectors $(\objectPoint{k} - \objectCentroid{k})$ and $(\objectPoint{l} -
\objectPoint{k})$ based on all points $\objectPoint{k}\in\objectPointCloud{k}$
and $\objectPoint{l}\in\objectPointCloud{l}$, i.e.,
\begin{equation}
\label{eqn:angleA}
\angA = \arccos\left(\frac{(\objectPoint{k} - \objectCentroid{k})\transpose (\objectPoint{l} -
\objectPoint{k})}{\|\objectPoint{k} - \objectCentroid{k}\|_2\quad\|\objectPoint{l} -
\objectPoint{k}\|_2}\right),
\end{equation}
where $\objectCentroid{k}$ is the centroid of the reference object $\object_k$, see~\figref{fig:pipas}.

For the same relative pose $\relativePose{l}{k}$, $\histAngA$
provides a unique signature of the scene that is invariant to its translation
or rotation in the global reference frame. 
However, in the context
of everyday manipulation tasks, it is typically useful to also reason about spatial constraints with
respect to the world frame, e.g., a supporting surface such
as a table. For example, this enables the robot to disambiguate scenes 
in which the two objects are on top of each other from those in which they are next to each
other for the same $\relativePose{l}{k}$. 

We achieve this disambiguation using the second histogram
$\histAngB$, which is a distribution over the angle $\angB$ around the vector
$(\objectPoint{k}-\objectCentroid{k})$, see~\figref{fig:pipas}. We take this
as the angle between two planes. The first plane is defined by the two
vectors $(\objectPoint{k}-\objectCentroid{k})$ and $\gravity$, whereas the
second is defined by the two vectors $(\objectPoint{k}-\objectCentroid{k})$ and 
$(\objectPoint{l}-\objectPoint{k})$. We compute $\angB$ as the angle between
the respective normal vectors $\normal_1$ and $\normal_2$ of those planes, i.e.,
\begin{align}
\normal_1 = \frac{(\objectPoint{k}-\objectCentroid{k}) \times
\gravity}{\|(\objectPoint{k}-\objectCentroid{k}) \times \gravity\|_2}&,
\normal_2 = \frac{(\objectPoint{k}-\objectCentroid{k}) \times
(\objectPoint{l}-\objectPoint{k})}{\|(\objectPoint{k}-\objectCentroid{k})
\times (\objectPoint{l}-\objectPoint{k})\|_2},\notag\\
\angB &= \arccos(\normal_1\transpose \normal_2) \label{eqn:angleB}.
\end{align}

We populate $\histAngB$ by computing $\angB$ using all surface points
$\objectPoint{k}\in\objectPointCloud{k}$ and
$\objectPoint{l}\in\objectPointCloud{l}$. As the direction of
$\gravity$ is fixed, rotating the scene while maintaining $\relativePose{l}{k}$ results in changes in
$\angB$, i.e., the discriminative behavior we seek. On the other hand,
$\histAngA$ and $\histAngB$ are invariant to translations or rotations around 
$\gravity$.

Whereas $\histAngA$ and $\histAngB$ encode the relation with respect
to the relative rotation between the two objects, we encode the desired
distance between them using the third histogram $\histDist$.
We compute $\histDist$ as a distribution over the Euclidean distance
$\|\objectPoint{k}-\objectPoint{l}\|_2$ between points $\objectPoint{k}\in\objectPointCloud{k}$ and
$\objectPoint{l}\in\objectPointCloud{l}$. Rather than doing so using all
$|\objectPointCloud{k}|\,|\objectPointCloud{l}|$ pairs
of points $\{\objectPoint{k}, \objectPoint{l}\}$, we consider the subset of
pairs with the smallest 10\% distances over all pairs, as this is
indicative of how close the two objects are and is less sensitive to
differences in object sizes.

We discretize both $\histAngA$ and $\histAngB$ with a bin
resolution of 20\,deg, and discretize $\histDist$ with a
resolution of 6\,cm. We normalize all histograms using the number of points used to 
compute them such that $\featureFunction(s)$ is independent of the object point cloud densities and object sizes. The resulting descriptor $\featureFunction(s)=\relationFeatureVector$ has 39 dimensions. To speed up computation, besides considering only surface points, we also downsample the point clouds. 
\section{Distance Metric Learning}
\label{sec:metricLearning}
In this section, we discuss how we learn a metric $\dist$ that models the
            similarities between relations given the feature
            representation $\relationFeatureVector = \featureFunction(\scene)$
            above (see~\secref{sec:distanceProblem}).
            For this, we leverage a popular metric learning technique originally 
            introduced to improve the performance of $k$-NN classification:
            large margin nearest neighbor (\lmnn)~\cite{weinberger2009distance}. 
                
            We follow the terminology of 
            \citeauthor{weinberger2009distance} and define the set of \emph{target 
            neighbors} $\setOfFriends{i}$ for an example 
            $\relationFeatureVector_i$ as the
       $k$ nearest neighbors of $\relationFeatureVector_i$ that are
       labeled as similar, i.e. $\simValueRowCol{i}{j} = 1$ for
       $\relationFeatureVector_j\in\setOfFriends{j}$.
     These target neighbors define a region around $\relationFeatureVector_i$. 
     We refer to all examples $\relationFeatureVector_k$ within this
       region that are not similar to $\relationFeatureVector_i$
       (i.e., $\simValueRowCol{i}{k}=0$) as \emph{imposters} $\setOfImposters{i}$.
       The original \lmnn~formulation identifies $\setOfFriends{i}$
       and $\setOfImposters{i}$ by assuming training data that is labeled with
       pre-specified classes. In our context, we achieve this based on the similarity
       labels $\simValue$ without requiring class labels to be specified by the
       teacher.  

In the general form, \lmnn~learns a metric $\distPhi$ parametrized by
$\distParams$ by minimizing a loss function with two objectives: \emph{i)} for
each training relation $\relationFeatureVector_i$, pull target neighbors
$\relationFeatureVector_j\in\setOfFriends{i}$ close, and \emph{ii)} push
imposters $\relationFeatureVector_k\in\setOfImposters{i}$ away such that they are further than
target neighbors $\relationFeatureVector_j$ by at least a large margin
$\lmnnMargin$ (see \cite{weinberger2009distance}), i.e., 
\begin{equation}
\label{eqn:lmnnOptimization}
\begin{aligned}
        &\minimize_{\distParams}
        \sum_{\substack{\relationFeatureVector_i\in\demos,\\
        \relationFeatureVector_j\in\setOfFriends{i}}}
        \overbrace{\distPhi(\relationFeatureVector_i,\relationFeatureVector_j)^2}^{\text{pull
        a similar neighbor $\relationFeatureVector_j$ close}} +\\
        &\pushPullConst \sum_{\relationFeatureVector_k\in\setOfImposters{i}}
        \underbrace{\big[\lmnnMargin +
                \distPhi(\relationFeatureVector_i,\relationFeatureVector_j)^2 -
\distPhi(\relationFeatureVector_i,\relationFeatureVector_k)^2\big]_+}_\text{push
a dissimilar neighbor $\relationFeatureVector_k$ further than $\relationFeatureVector_j$
by at least $\lmnnMargin$},
\end{aligned}
\end{equation}
where $[d]_+ = \max(0, d)$ is the hinge loss and $\pushPullConst$ is a
constant that controls the trade-off between the two objectives.


In this work, we consider three \lmnn-based methods for learning a $\distPhi$
parametrized by $\distParams$. The linear \lmnn~case learns a generalized
Euclidean (Mahalanobis) distance by parametrizing $\distPhi$ using a linear
mapping~$\metricMatrixDecomposed\in\realNumbers^{\relationFeatureVectorDim\times
\relationFeatureVectorDim}$, i.e., $\distParams(\relationFeatureVector) =
\metricMatrixDecomposed \relationFeatureVector$, see~\cite{weinberger2009distance}. \chiSqlmnn~also
learns a linear mapping but uses the $\chiSq$
distance instead of the Euclidean distance, see~\cite{kedem2012non}. Finally,
gradient-boosted~\lmnn~(\gblmnn) models arbitrary non-linear mappings
$\distParams(\relationFeatureVector)$ of the input space using gradient-boosted
regression trees, see~\cite{kedem2012non}.

\section{Reproducing a New Relation from a Few Demonstrations}
\label{sec:interactiveImitation}
In this section, we present our approach for  
imitating a new relation from a small number of teacher demonstrations (\secref{sec:imitationProblem}).
We assume that the robot is already equipped with a set of relation scenes
$\demos=\{\sceneNthDemo{1}, \dots, \sceneNthDemo{N}\}$ and a (partially-filled) 
$N\times N$ matrix $\simMatrix$ consisting of their similarity
labels as in~\secref{sec:distanceProblem}. These are either provided by an
expert beforehand, or are accumulated by the robot when learning previous
relations over time. Using $\demos$ and $\simMatrix$, we assume the robot has
already learned a prior distance metric $\priorMetric$ parametrized by
$\priorDistParams$ as
described in~\secref{sec:metricLearning}. This is done offline and without knowledge 
of the new relation.

We now consider a teacher providing the robot with a small set of demonstrations
$\newRelationDemos$ of size $N'$ for a new,
arbitrary relation. The teacher can use different pairs of objects, 
such that all scenes in $\newRelationDemos$ are equally valid 
ways of achieving this relation, i.e., $\simValueRowCol{i}{j} = 1$ for all
$\sceneNthDemo{i},\sceneNthDemo{j}\in\newRelationDemos$. Given two 
objects $\object_k$ and $\object_l$ and their respective models
$\objectPointCloud{k}$ and $\objectPointCloud{l}$, our goal is to 
compute a pose $\relativePoseOptimal{l}{k}$ such that the resulting scene 
$\testScene = \langle\objectPointCloud{k}, \objectPointCloud{l},
\,\relativePoseOptimal{l}{k}\rangle$ corresponds to the intention of the
teacher for the new relation, i.e., minimizing $\lossFunction$ in~\eqref{eqn:imitationProblem}.

Given a metric $\onlineMetric$, there are different ways to
model $\lossFunction$ to express the distance between the features
$\testSceneFeature=\featureFunction(\testScene)$ of the test scene and the features
$\setOfNewRelationFeautres$ of the demonstrations $\newRelationDemos$. In
general, as $|\setOfNewRelationFeautres|\geq1$, \eqref{eqn:imitationProblem} is
a multi-objective optimization problem seeking to minimize the distance between $\testSceneFeature$ and
all $\newRelationFeautreExample\in\setOfNewRelationFeautres$.
In such settings, it is
typically challenging to satisfy all objectives. Minimizing the 
(mean) distance to $\newRelationDemos$ can thus 
lead to sub-optimal solutions that ``average'' the demonstrated scenes.
Instead, we consider each demonstration to represent a mode of the target relation and
seek the best solution with respect to any of them as follows:
\begin{equation}
\label{eqn:imitationObjectiveFunc}
    \lossFunction(\setOfNewRelationFeautres,\testSceneFeature) :=
    \min\limits_{\newRelationFeautreExample\in\setOfNewRelationFeautres}
    \onlineMetric(\newRelationFeautreExample,\testSceneFeature).
\end{equation}

\subsection{Interactive Local Metric Learning}
\label{sec:imitationObjectiveFunc}
\begin{figure*}[t]
\centering
\includegraphics[width=0.8\linewidth]{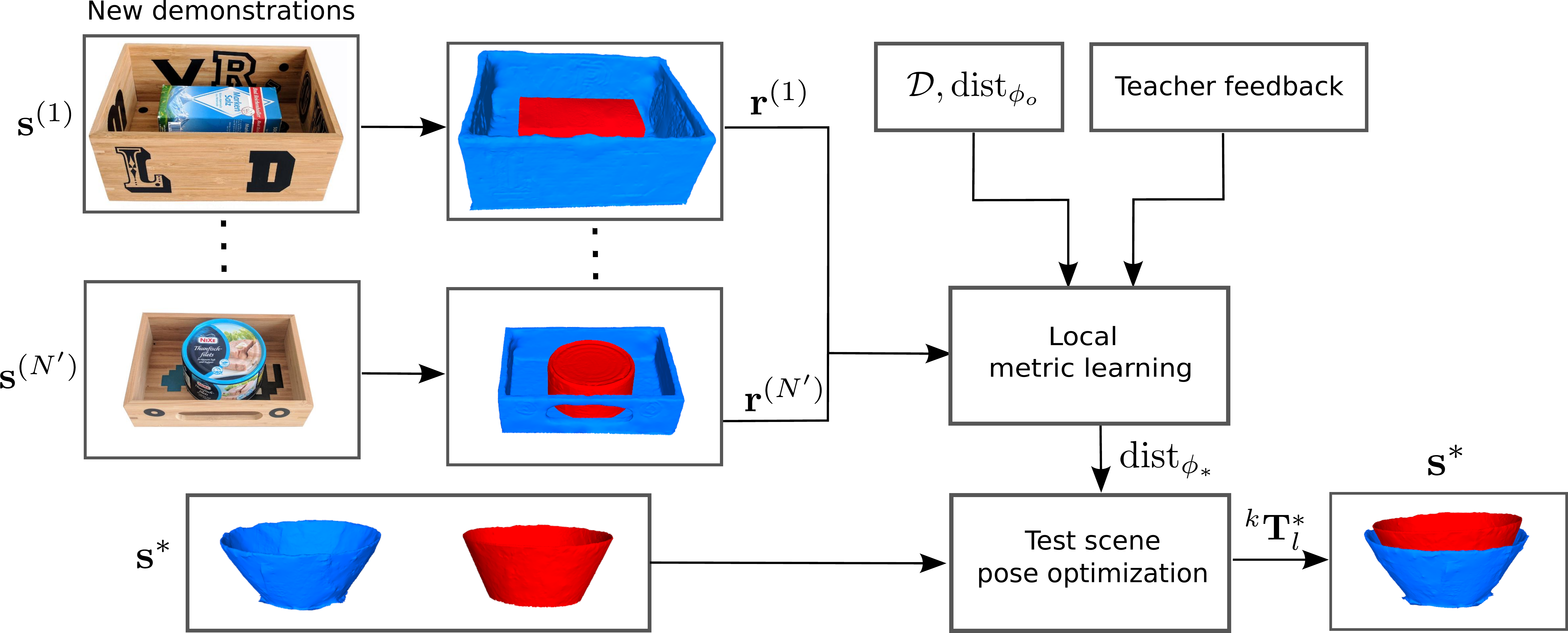}
\caption{\label{fig:approachOverview}
Overview of our interactive approach for learning to reproduce a new
relation. Given a small number of demonstrations $\sceneNthDemo{1}, \dots,
\sceneNthDemo{N'}$ by a teacher and two objects in the test scene $\testScene$, we aim to
compute a pose $\relativePoseOptimal{l}{k}$ transforming $\testScene$ in order to imitate
the demonstrated relation and generalize the intention of the teacher. Our
approach enables the robot to leverage its prior knowledge of
scenes $\demos$ representing other relations and the distances between them
based on a metric $\priorMetric$.}
\end{figure*}


One way to obtain $\onlineMetric$ in~\eqref{eqn:imitationObjectiveFunc} is to 
use the prior metric
$\priorMetric$. However, we learned this metric using a set of 
previous relations $\demos$ and their similarities. Therefore, it is not 
directly clear if $\priorMetric$ is able to generalize to novel relations shown 
by the teacher.

We answer this question using an interactive approach. For 
each demonstration in $\newRelationDemos$, we use $\priorMetric$
to retrieve the $\numSceneNN$ nearest neighbor examples corresponding to scenes from the 
database $\demos$. We query the teacher with these examples and ask her to
indicate whether they align with her intention ($\simValue=1$) for the new
relation or not ($\simValue=0$). In
our experiments, we achieved this by means of a graphical user
interface visualizing $\numSceneNN = 8$ nearest neighbors per query. 
Let $\nnScenes\subset\demos$ be the set of all nearest
neighbor scenes for $\newRelationDemos$. We measure the confidence in
the ability of $\priorMetric$ to generalize to the new relation as the ratio
$\nnRatioGood$ of scenes in $\nnScenes$ for which the teacher indicated a similarity 
to the new relation (i.e., $\simValue=1$).

$\nnRatioGood$ values larger than a threshold $\nnRatioThresh$ indicate that we are 
able to relate the new relation to ones the robot has seen in the past. Therefore, we
use $\priorMetric$ to compute \eqref{eqn:imitationObjectiveFunc}, i.e., $\onlineDistParams =
\priorDistParams$. We empirically set $\nnRatioThresh$ to 77$\%$ in our experiments. 
On the other hand, $\nnRatioGood<\nnRatioThresh$ indicates
that the new relation is far in feature space from (target neighbor) relations
in $\demos$. We address this by learning a new \emph{local} metric
$\onlineMetric$ using the set of scenes $\onlineMetricDemos = \newRelationDemos \cup \nnScenes$ and
labels $\simMatrix^*$ of size $N^*\times N^*$, where $N^* =
|\onlineMetricDemos|$.  This is a smaller problem (compared to learning the prior
metric $\priorMetric$) in which $\simMatrix^*$ is completely specified. 
We set the similarity $\simValueRowCol{i}{j}$ to 1 for all
$\sceneNthDemo{i},\sceneNthDemo{j}\in\newRelationDemos$.
For rows $i$ and columns $j$ corresponding to scenes 
$\sceneNthDemo{i}\in\newRelationDemos$ and
$\sceneNthDemo{j}\in\nnScenes$ (or vice versa), we set $\simValueRowCol{i}{j}$
to the similarity labels obtained from querying the teacher. We use the
transitivity property to set the similarity between $\sceneNthDemo{j},
\sceneNthDemo{k} \in\nnScenes$. For example, if the teacher labeled
$\simValueRowCol{i}{j}=0$ and $\simValueRowCol{i}{k}=0$, we set
$\simValueRowCol{j}{k}=1$.



Finally, we highlight two main advantages of leveraging the previous relations $\demos$
and prior metric $\priorMetric$.
Firstly, we enable the robot to consider
whether its previous knowledge is sufficient to reproduce the new relation or
not. Secondly, even for new relations that are significantly different
from previously-known ones, we are able to augment the teacher's demonstrations
$\newRelationDemos$ with additional training data $\nnScenes$ consisting of target 
neighbors and imposters retrieved from $\demos$ without requiring the teacher
to demonstrate them.

\subsection{Sample-Based Pose Optimization}
\label{sec:gridSearch}
Given the metric $\onlineMetric$ to model~\eqref{eqn:imitationObjectiveFunc}, we
present our approach for solving~\eqref{eqn:imitationProblem} to compute
$\relativePoseOptimal{l}{k}$ for reproducing the new relation using $\object_k$
and $\object_l$. In this work, we simplify this problem by assuming that the
reference object $\object_k$ is stationary and therefore only reason about
desirable poses of $\object_l$ relative to it. Due to the discretization in
computing our descriptor $\featureFunction$, we cannot rely on gradient-based
methods as our loss function is piecewise
constant.

We address this using a sample-based approach. We discretize the space of poses
by searching over a grid of translations $\transVecRel{l}{k}$ of $\object_l$
relative to $\object_k$. For each translation, we sample rotations 
$\rotMatRel{l}{k}$ uniformly.
We use the resulting $\relativePose{l}{k}$ to transform $\objectPointCloud{l}$ 
and compute $\lossFunction$ 
based on the corresponding feature value $\testSceneFeature$ of the scene. Whenever we find a new local minima during optimization, we
check for collisions between the two objects and reject infeasible solutions. Finally, 
we take $\relativePoseOptimal{l}{k}$ as the
feasible pose minimizing $\lossFunction$ over all sampled poses. We 
implemented this process efficiently by parallelizing the grid search over
translations.~\figref{fig:approachOverview} shows an overview of our method for
reproducing a new relation.





\section{Experimental Evaluation}
\label{sec:metricExperiments}

In this section, we present the experimental evaluation of our approach.
Through our experiments, we demonstrate the following: \emph{i)} our proposed
descriptor is able to capture different spatial relations and to generalize to
the shapes and sizes of the objects, \emph{ii)} using distance metric learning, we are
able to capture the similarities between scenes even for relations 
not encountered before, \emph{iii)} our interactive learning method enables
non-expert users to teach new relations based on small number of examples,
and \emph{iv)} we outperform several baselines that do not learn a metric based on
the similarities between scenes.


\subsection{Baselines}
\label{sec:metricBaselines}
In our experiments, we used three variants of \lmnn-based metrics: vanilla
(linear) \lmnn, \chiSqlmnn, and \gblmnn, which we learned as
in~\secref{sec:metricLearning}. We compared those learned metrics
to a variety of standard distance metrics. This includes the Euclidean, 
$\chi^2$, Bhattacharyya, and the correlation distances, as well as the Kullback-Leibler divergence (KL) and
the Jensen-Shannon divergence (JS).

\subsection{Dataset}
\label{sec:metricRecordingData}
\begin{figure}[h]
  \centering
\includegraphics[width=0.5\textwidth]{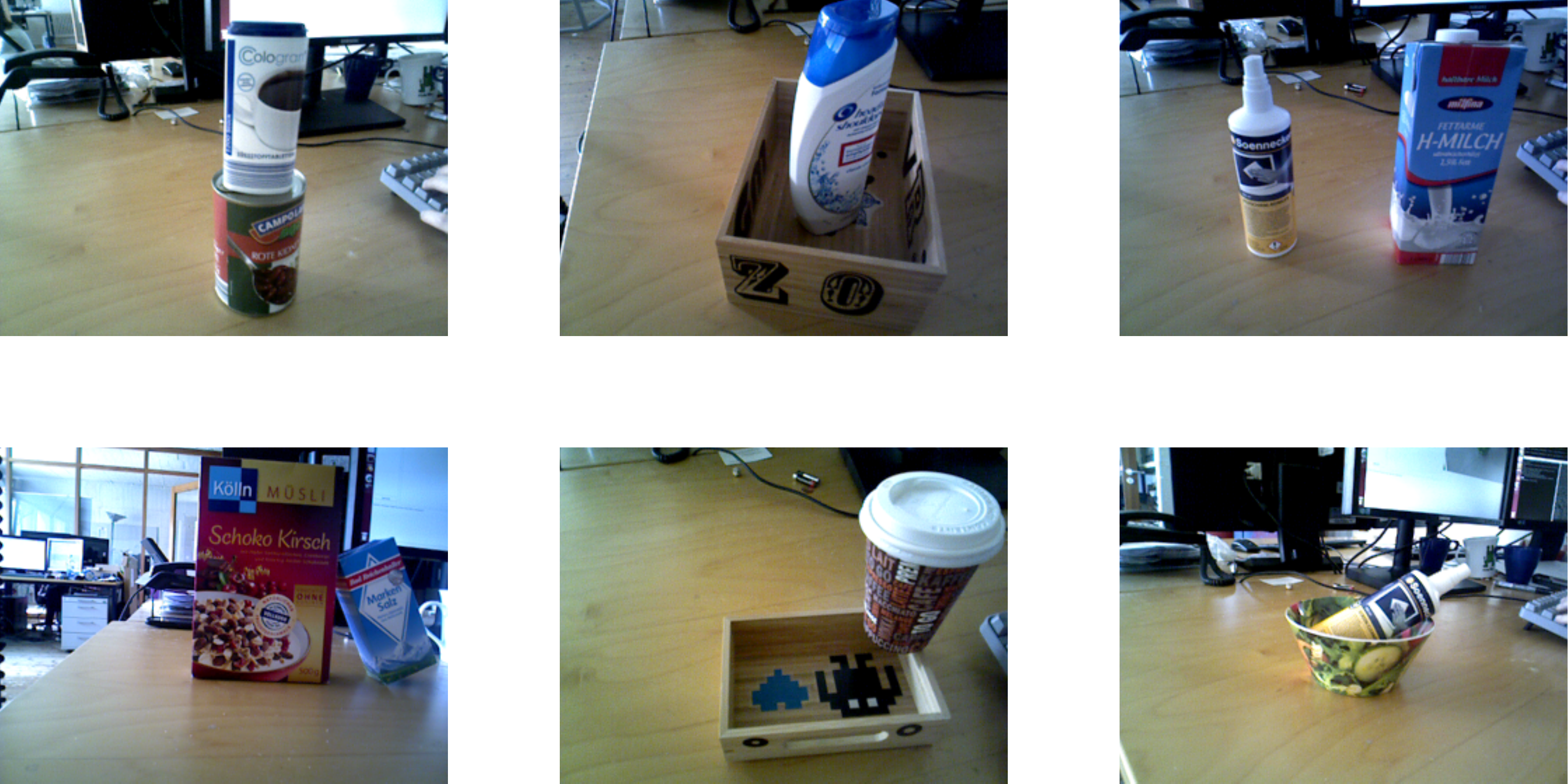}
  \caption{\label{fig:six_classes} 
          Examples of the training scenes we recorded for our evaluation. We
          recorded a set of 546 scenes and manually provided ground truth
          labels for their similarities.}
\end{figure}

We recorded 3D models of 26 household objects and used SimTrack to detect
them and compute their poses in a scene using a Kinect camera~\cite{pauwels2015simtrack}.
Using this setup, we recorded a set of demonstrations $\demos$ consisting of
546 scenes, see~\figref{fig:six_classes} for examples. For 
the purpose of evaluation, we manually labeled the similarities $\simMatrix$ between all
scenes. 

\subsection{Nearest Neighbor Classification}
\label{sec:metricNNExperiment}
In this experiment, we evaluated the ability of distance metric learning to
relate scenes based on the similarities of their relations. We formulated this
as a $k$-NN classification problem, with $k=5$, and evaluated using 15 random splits. 
For each split we used 75\% of the data for the training set and 25\% for the
test set. We considered a success if at least 3 out of 5 of the retrieved
nearest neighbors were similar to the test example.
\begin{table}[h]
    \centering
    \begin{tabular}{c c  }
  Method & Accuracy(\%)               \\
  \hline
  \hline
  Euclidean & $82.32 \pm 2.56$         \\
  KL    &  $82.61 \pm 3.10$        \\
  Correlation & $ 82.66 \pm 2.43$      \\ 
  $\chi^2$ & $82.81 \pm 3.20$            \\ 
  Bhattacharyya & $83.26  \pm 3.15$     \\
  JS  & $83.30     \pm 3.14$        \\
  $\chi^2$-LMNN   &$ 86.46 \pm 2.84$    \\ 
  LMNN & $86.52    \pm 1.98$          \\         
  GB-LMNN &  $\mathbf{87.60 \pm 1.94}$           \\ 
  
    \end{tabular}
    \caption{\label{tab:knn}Performance of different methods for retrieving at least
    3 out of 5 target neighbors of scenes, averaged over 15 random splits.} 
    \label{tab:knnresults}
\end{table} 

\begin{figure}[t]
  \centering
\includegraphics[width=0.5\textwidth]{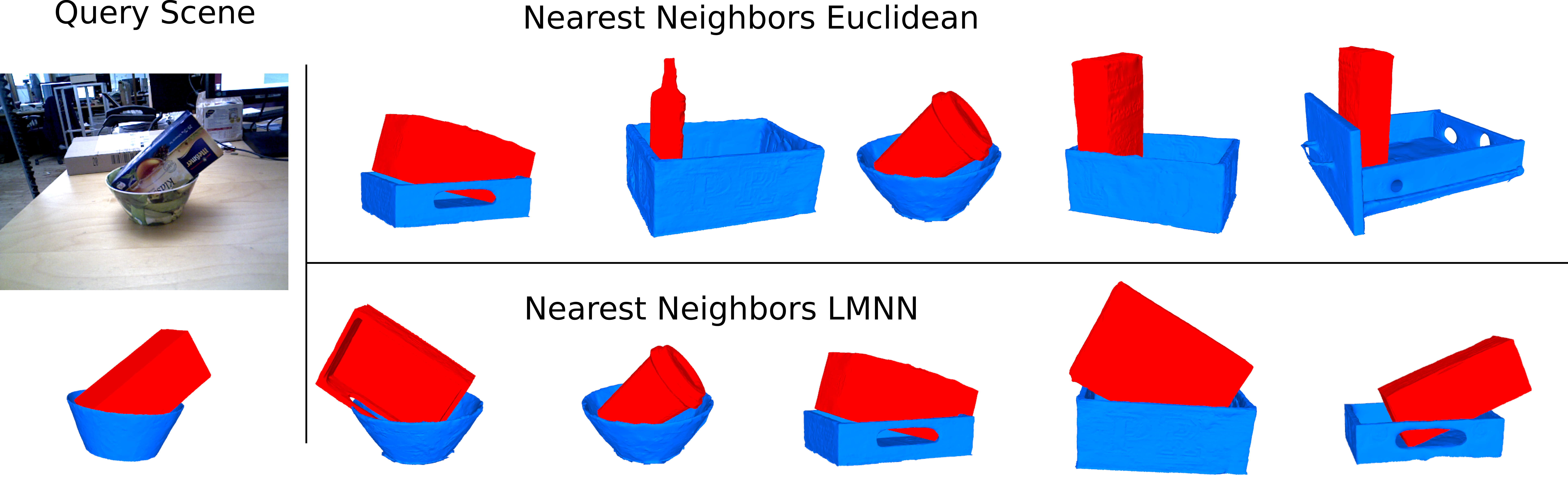}
\caption{\label{fig:exampleNNQuery} Example of the 5 nearest neighbors of
        a query scene using Euclidean distance (top) and~\lmnn~(bottom). \lmnn~better
captures the distances between relations.} 
  \end{figure}

The results are shown in~\tabref{tab:knn}. \lmnn-based metrics outperform the
baselines, i.e., the learned metrics can better capture the distances between
scenes. We achieved the highest success rate of $87.6\%$ using~\gblmnn. Note
that by directly computing the Euclidean distance in the original feature
space, we are able to achieve a success rate of $82.32\%$. This demonstrates that 
our proposed feature descriptor is suitable for encoding arbitrary spatial relations. 
\figref{fig:exampleNNQuery} shows a qualitative example of the nearest neighbors of a 
test scene using both the Euclidean distance and \lmnn.


\subsection{Distance to New Relations}
\label{sec:metricNewRelationsExp}
\begin{figure}[t]
  \centering
\includegraphics[width=0.45\textwidth]{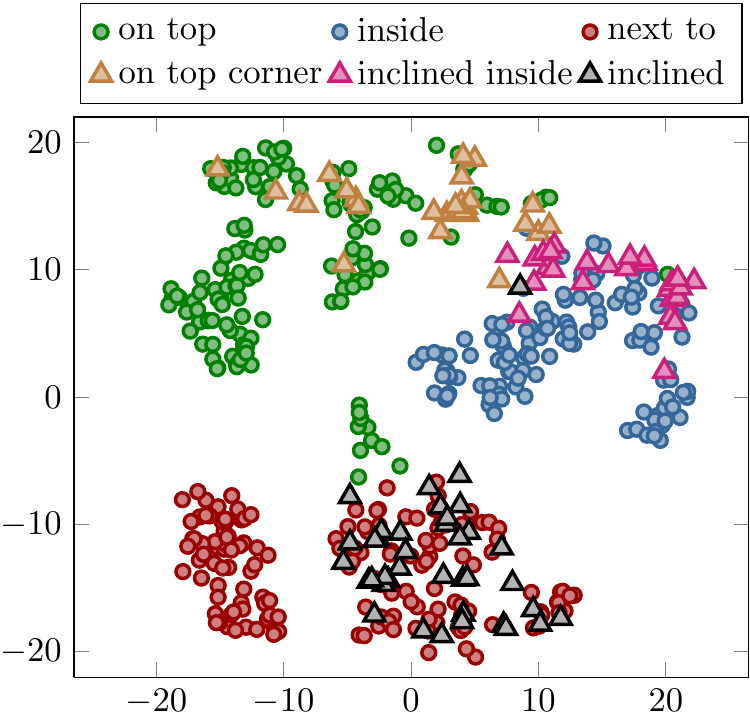}
  \caption{\label{fig:tsne} t-SNE visualization of six relations mapped using \lmnn. 
          We trained this metric using three of the
          relations only, yet it is able to capture the semantic similarity between those
          relations and the test relations.}
\end{figure}

In this qualitative experiment, we investigated the ability of a learned
metric to capture the similarities between relations that were not used for
training. We trained \lmnn~with data from three relations
(upper row of \figref{fig:six_classes}), which can be semantically described as
``on top'', ``inside'', and ``next to''. We used the learned metric to map all six
relations in~\figref{fig:six_classes} to the new space and visualized the data using
t-SNE, a popular non-linear embedding technique for visualizing high
dimensional data~\cite{maaten2008visualizing}. We show this
in~\figref{fig:tsne}. This qualitatively illustrates the separation between the three
relations used for training the metric. Moreover, the metric is able to capture
the semantic similarity between the relations used for training and the new
ones, which we denote by ``inclined'', ``on top corner'' and ``inclined
inside'' (bottom row of~\figref{fig:six_classes}).

%

\subsection{Generalizing a Relation to New Objects}
\label{sec:metricNewObjectsExp}

In this  experiment, we evaluated our approach for reproducing a demonstrated
relation using two new objects, see~\secref{sec:imitationProblem}. 
We recorded 30 demonstrations for each of five new relations.
We then selected two new objects that were not used in the demonstrations and
evaluated our method's ability to generalize each of the five relations to the
new objects. 
In each case, we provided our method with $|\newRelationDemos|=5$ examples.
Using those examples, we retrieved nearest neighbor queries $\nnScenes$ from $\demos$ as
described in~\secref{sec:interactiveImitation}. 
As we aimed for a quantitative evaluation in this experiment,
we implemented an ``oracle'' that provides the binary labels for the queries
automatically, and used this to learn a local metric as in~\secref{sec:imitationObjectiveFunc}. 

For evaluation, we provided our method with a set of 75 poses $\relativePose{l}{k}$ between the
new objects. Only 15 of them are correct ways of
imitating the relation in question, whereas the rest represent other relations.
For each pose, we computed $\testSceneFeature$ of the test scene and the corresponding
$\lossFunction$ (\eqref{eqn:imitationObjectiveFunc}). We sorted the poses
according to $\lossFunction$. Ideally, the correct poses should be in the top 15 positions. 
We evaluated this using the mean average precision.

\begin{figure}[t]
  \centering
\includegraphics[width=0.5\textwidth]{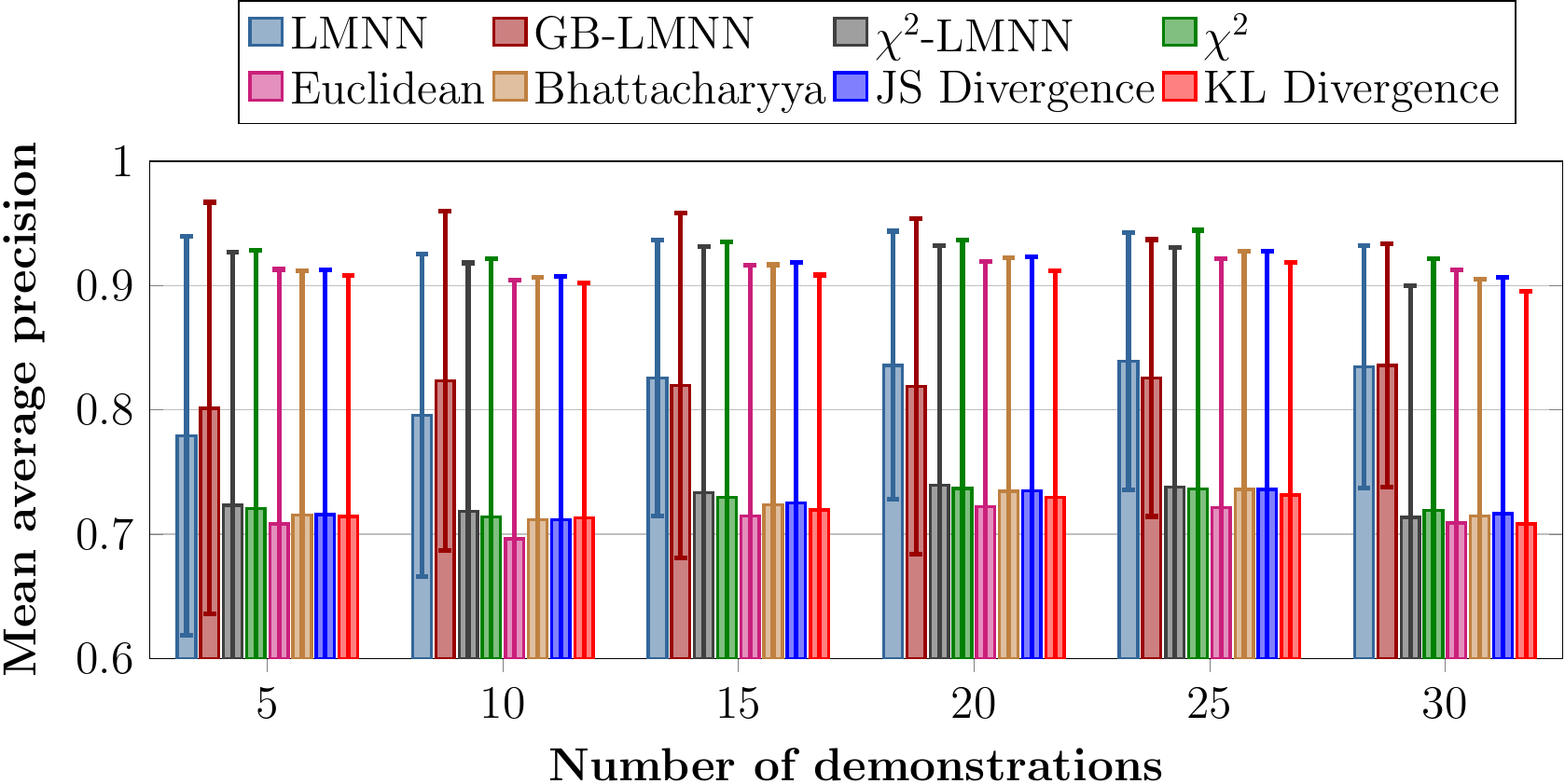}
\caption{\label{fig:generalization} Results for generalizing a relation to two
        new objects. \gblmnn~and \lmnn~outperform the other metrics in
identifying the correct ways of reproducing a relation with two new objects.
Our approach enables leveraging previous demonstrations when querying the
teacher to learn a new relation. In each case, we evaluated the generalization given five new demonstrations, which
we then added to the set of prior demonstrations $\demos$. 
The x-axis shows the cumulative number of demonstrations over time.}
\end{figure}

After each such test, we added the five demonstrations $\newRelationDemos$ to
the database $\demos$ and extended $\simMatrix$ with the new
labels from the nearest neighbor queries. We used this to re-learn the prior metric
$\priorMetric$. 
We did this six times, each with five new demonstrations, until the 30
demonstrations have been used. We repeated the whole experiment 50 times using
different random orders of providing five demonstrations. We report the results in~\figref{fig:generalization}
averaged over all runs and five relations.


The metrics we learned with \gblmnn~and \lmnn~outperform the other metrics in
their ability to identify the correct ways of reproducing a relation with two
new objects. Although in each case we compute~\eqref{eqn:imitationObjectiveFunc} 
based on five new demonstrations only,
our approach enables those metrics to re-use demonstrations added to $\demos$
from the previous tests to learn the local metric. Accordingly, 
\gblmnn~achieves a mean average precision of 0.82 after having seen at least
five demonstrations in the past. This demonstrates the ability of our approach
to use previous demonstrations when generalizing a new relation in a lifelong
learning manner.


\subsection{Interactive Learning of a New Relation}
\label{sec:metricInteractiveExp}

We conducted a small survey to evaluate our approach for learning a new
relation interactively. We recorded 250 scenes consisting of 50
relations demonstrated by nine different teachers. Each teacher provided five
demonstrations per relation using different objects they chose. We used only
three of those demonstrations to learn each relation. In each case, we queried
the teacher with nearest neighbor examples from $\demos$ and used the result to
generalize the three demonstrations $\newRelationDemos$ and reproduce one of the scenes we left out
from the training as in~\secref{sec:interactiveImitation}. We computed the best
pose for reproducing the relation using the sample-based approach in~\secref{sec:gridSearch}.

We showed the reproduced relations to their corresponding teachers in a 3D
visualization environment and asked
them to score the quality of the result with 0, 0.5, or 1, where 0 represents
unsuccessful and 1 represents successful. The teachers scored results from different
baselines shown in random order. 

\begin{figure}[t]
  \centering
\includegraphics[width=0.5\textwidth]{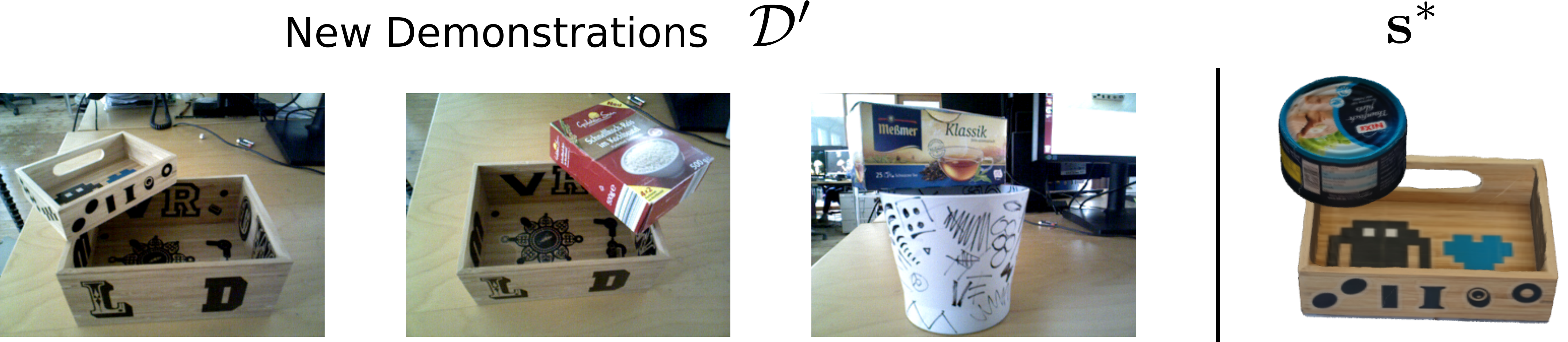}
  \caption{\label{fig:good_reconstruction1} Left: three demonstrations of a
          relation by a teacher in our survey. Right: the generalized relation with two new objects using our approach.}
\end{figure}

We show the mean scores in~\tabref{tab:reconstructionScores}. The reproduced
scenes using both \lmnn~and \gblmnn~metrics were judged to be the best by the
teachers, achieving an average score of 0.72 and 0.71 respectively.
\figref{fig:good_reconstruction1}~illustrates one such generalization from our
experiments. The results confirm that our approach enables non-expert users to 
teach arbitrary spatial relations to a robot from a small number of examples. 


\begin{table}[h]
    \centering
    \begin{tabular}{c c }
  Method &   Score          \\
  \hline
  \hline
  Euclidean &  0.49        \\
  Bhattacharyya & 0.54  \\
  KL    &  0.56       \\
  JS  & 0.56       \\
  $\chi^2$ & 0.60          \\  
  $\chi^2$-LMNN   & 0.63    \\ 
  LMNN & \textbf{0.72}        \\         
  GB-LMNN &  0.71         \\ 
  
    \end{tabular}
    \caption{Mean scores for reproducing 50 relations from
    nine different teachers. For each relation, the teacher scored the
    result with 0 (unsuccessful), 0.5, or 1 (successful).}
    \label{tab:reconstructionScores}
\end{table}

As a final qualitative evaluation, 
we carried out a small survey in which we asked six participants to judge 
the quality of scenes generated by our sample-based approach in~\secref{sec:gridSearch}. 
For this, we selected 30 scenes computed by \lmnn, and which were scored with
either 0.5 or 1 in the experiment above. We asked the six participants to judge
whether the relations in those scenes were demonstrated by a human or generated
by a computer (whereas in fact they were all computed using our approach).
Despite the fact that we do not consider physics checks (e.g., scene stability)
or optimize computed scenes to make them more realistic, 63.05\% of the scenes were 
thought to have been produced by a human.




\section{Conclusion}
In this paper, we presented a novel approach to the problem of learning
pairwise spatial relations and generalizing them to different objects. Our
method is based on distance metric learning and enables a robot to reason about
the similarity between scenes with respect to the relations they represent. 
To encode a relation, we introduced a novel descriptor based on the geometries of 
the objects. By learning a distance metric using this
representation, our method is able to reproduce a new relation from a small
number of teacher demonstrations by reasoning about its similarity to
previously-encountered ones. In this way, our approach allows for a lifelong
learning scenario by continuously leveraging its
prior knowledge about relations to bootstrap imitating new ones. We evaluated
our approach extensively using real-world data we gathered from non-expert
teachers. Our results demonstrate the effectiveness of our approach in
reasoning about the similarity between relations and its ability to reproduce
arbitrary relations to new objects by learning interactively from a teacher. In the future, we plan to extend the approach to partial observations of point cloud data.
\label{sec:conclusion}


\section*{Acknowlegments}
This work has partly been supported by the German Research
Foundation under research unit FOR 1513 (HYBRIS) and grant number
EXC 1086. We thank Gian Diego Tipaldi, Marc Toussaint, and Oliver Kroemer for the valuable discussions during the early stages of this work.


\footnotesize
\bibliographystyle{abbrvnat}
\bibliography{references}

\begin{thebibliography}{27}
\providecommand{\natexlab}[1]{#1}
\providecommand{\url}[1]{\texttt{#1}}
\expandafter\ifx\csname urlstyle\endcsname\relax
  \providecommand{\doi}[1]{doi: #1}\else
  \providecommand{\doi}{doi: \begingroup \urlstyle{rm}\Url}\fi

\bibitem[Abdo et~al.(2013)Abdo, Kretzschmar, Spinello, and
  Stachniss]{abdo2013learning}
N.~Abdo, H.~Kretzschmar, L.~Spinello, and C.~Stachniss.
\newblock Learning manipulation actions from a few demonstrations.
\newblock In \emph{Int.~Conf.~on Robotics \& Automation (ICRA)}, 2013.

\bibitem[Ahmadzadeh et~al.(2015)Ahmadzadeh, Paikan, Mastrogiovanni, Natale,
  Kormushev, and Caldwell]{ahmadzadeh2015learning}
S.~R. Ahmadzadeh, A.~Paikan, F.~Mastrogiovanni, L.~Natale, P.~Kormushev, and
  D.~G. Caldwell.
\newblock Learning symbolic representations of actions from human
  demonstrations.
\newblock In \emph{Robotics and Automation (ICRA), 2015 IEEE International
  Conference on}, pages 3801--3808. IEEE, 2015.

\bibitem[Aydemir and Jensfelt(2012)]{aydemir2012exploiting}
A.~Aydemir and P.~Jensfelt.
\newblock Exploiting and modeling local 3d structure for predicting object
  locations.
\newblock In \emph{Int.~Conf.~on Intelligent Robots and Systems (IROS)}, 2012.

\bibitem[Beetz et~al.(2010)Beetz, M{\"o}senlechner, and Tenorth]{beetz2010cram}
M.~Beetz, L.~M{\"o}senlechner, and M.~Tenorth.
\newblock {CRAM-A} cognitive robot abstract machine for everyday manipulation
  in human environments.
\newblock In \emph{Int.~Conf.~on Intelligent Robots and Systems (IROS)}, 2010.

\bibitem[Fichtl et~al.(2014)Fichtl, McManus, Mustafa, Kraft, Kr{\"u}ger, and
  Guerin]{fichtl2014learning}
S.~Fichtl, A.~McManus, W.~Mustafa, D.~Kraft, N.~Kr{\"u}ger, and F.~Guerin.
\newblock Learning spatial relationships from 3d vision using histograms.
\newblock In \emph{2014 IEEE International Conference on Robotics and
  Automation (ICRA)}, pages 501--508. IEEE, 2014.

\bibitem[Guadarrama et~al.(2013)Guadarrama, Riano, Golland, Go, Jia, Klein,
  Abbeel, Darrell, et~al.]{guadarrama2013grounding}
S.~Guadarrama, L.~Riano, D.~Golland, D.~Go, Y.~Jia, D.~Klein, P.~Abbeel,
  T.~Darrell, et~al.
\newblock Grounding spatial relations for human-robot interaction.
\newblock In \emph{2013 IEEE/RSJ International Conference on Intelligent Robots
  and Systems}, pages 1640--1647. IEEE, 2013.

\bibitem[Guillaumin et~al.(2009)Guillaumin, Verbeek, and
  Schmid]{guillaumin2009you}
M.~Guillaumin, J.~Verbeek, and C.~Schmid.
\newblock Is that you? metric learning approaches for face identification.
\newblock In \emph{2009 IEEE 12th International Conference on Computer Vision},
  pages 498--505. IEEE, 2009.

\bibitem[H{\"o}fer and Brock(2016)]{hofer2016coupled}
S.~H{\"o}fer and O.~Brock.
\newblock Coupled learning of action parameters and forward models for
  manipulation.
\newblock In \emph{Int.~Conf.~on Intelligent Robots and Systems (IROS)}, 2016.

\bibitem[Hoi et~al.(2008)Hoi, Liu, and Chang]{hoi2008semi}
S.~C. Hoi, W.~Liu, and S.-F. Chang.
\newblock Semi-supervised distance metric learning for collaborative image
  retrieval.
\newblock In \emph{Computer Vision and Pattern Recognition, 2008. CVPR 2008.
  IEEE Conference on}, pages 1--7. IEEE, 2008.

\bibitem[Jetchev et~al.(2013)Jetchev, Lang, and Toussaint]{jetchev2013learning}
N.~Jetchev, T.~Lang, and M.~Toussaint.
\newblock Learning grounded relational symbols from continuous data for
  abstract reasoning.
\newblock ICRA Workshop on Autonomous Learning, 2013.

\bibitem[Jiang et~al.(2012)Jiang, Lim, Zheng, and Saxena]{jiang2012learning}
Y.~Jiang, M.~Lim, C.~Zheng, and A.~Saxena.
\newblock Learning to place new objects in a scene.
\newblock \emph{Int.~J.~of Robotics Research (IJRR)}, 2012.

\bibitem[Kaelbling and Lozano-P{\'e}rez(2013)]{kaelbling2013integrated}
L.~P. Kaelbling and T.~Lozano-P{\'e}rez.
\newblock Integrated task and motion planning in belief space.
\newblock \emph{Int.~J.~of Robotics Research (IJRR)}, 32\penalty0
  (9-10):\penalty0 1194--1227, 2013.

\bibitem[Kedem et~al.(2012)Kedem, Tyree, Sha, Lanckriet, and
  Weinberger]{kedem2012non}
D.~Kedem, S.~Tyree, F.~Sha, G.~R. Lanckriet, and K.~Q. Weinberger.
\newblock Non-linear metric learning.
\newblock In \emph{Advances in Neural Information Processing Systems}, pages
  2573--2581, 2012.

\bibitem[Konidaris et~al.(2014)Konidaris, Kaelbling, and
  Lozano-Perez]{konidaris2014constructing}
G.~Konidaris, L.~P. Kaelbling, and T.~Lozano-Perez.
\newblock Constructing symbolic representations for high-level planning.
\newblock In \emph{National Conf.~on Artificial Intelligence (AAAI)}, 2014.

\bibitem[Kroemer and Peters(2014)]{kroemer2014predicting}
O.~Kroemer and J.~Peters.
\newblock Predicting object interactions from contact distributions.
\newblock In \emph{2014 IEEE/RSJ International Conference on Intelligent Robots
  and Systems}, pages 3361--3367. IEEE, 2014.

\bibitem[Kulick et~al.(2013)Kulick, Toussaint, Lang, and
  Lopes]{kulick2013active}
J.~Kulick, M.~Toussaint, T.~Lang, and M.~Lopes.
\newblock Active learning for teaching a robot grounded relational symbols.
\newblock In \emph{IJCAI}, 2013.

\bibitem[Lai et~al.(2011)Lai, Bo, Ren, and Fox]{lai2011sparse}
K.~Lai, L.~Bo, X.~Ren, and D.~Fox.
\newblock Sparse distance learning for object recognition combining rgb and
  depth information.
\newblock In \emph{Robotics and Automation (ICRA), 2011 IEEE International
  Conference on}, pages 4007--4013. IEEE, 2011.

\bibitem[Lazebnik et~al.(2005)Lazebnik, Schmid, and Ponce]{rift}
S.~Lazebnik, C.~Schmid, and J.~Ponce.
\newblock A sparse texture representation using local affine regions.
\newblock \emph{IEEE Transactions on Pattern Analysis and Machine
  Intelligence}, 27\penalty0 (8):\penalty0 1265--1278, 2005.

\bibitem[Maaten and Hinton(2008)]{maaten2008visualizing}
L.~v.~d. Maaten and G.~Hinton.
\newblock Visualizing data using t-sne.
\newblock \emph{Journal of Machine Learning Research}, 9\penalty0
  (Nov):\penalty0 2579--2605, 2008.

\bibitem[Pasula et~al.(2007)Pasula, Zettlemoyer, and
  Kaelbling]{pasula2007learning}
H.~M. Pasula, L.~S. Zettlemoyer, and L.~P. Kaelbling.
\newblock Learning symbolic models of stochastic domains.
\newblock \emph{Journal of Artificial Intelligence Research}, 29:\penalty0
  309--352, 2007.

\bibitem[Pauwels and Kragic(2015)]{pauwels2015simtrack}
K.~Pauwels and D.~Kragic.
\newblock Simtrack: A simulation-based framework for scalable real-time object
  pose detection and tracking.
\newblock In \emph{Intelligent Robots and Systems (IROS), 2015 IEEE/RSJ
  International Conference on}, pages 1300--1307. IEEE, 2015.

\bibitem[Paxton et~al.(2016)Paxton, Jonathan, Kobilarov, and
  Hager]{paxton2016want}
C.~Paxton, F.~Jonathan, M.~Kobilarov, and G.~D. Hager.
\newblock Do what i want, not what i did: Imitation of skills by planning
  sequences of actions.
\newblock In \emph{Int.~Conf.~on Intelligent Robots and Systems (IROS)}, 2016.

\bibitem[Rosman and Ramamoorthy(2011)]{rosman2011learning}
B.~Rosman and S.~Ramamoorthy.
\newblock Learning spatial relationships between objects.
\newblock \emph{The International Journal of Robotics Research}, 30\penalty0
  (11):\penalty0 1328--1342, 2011.

\bibitem[Shahid et~al.(2016)Shahid, Naseer, and Burgard]{shahid16rss_ws}
M.~Shahid, T.~Naseer, and W.~Burgard.
\newblock Dtlc: Deeply trained loop closure detections for lifelong visual
  slam.
\newblock In \emph{Visual Place Recognition: What is it Good For? Workshop at
  the Robotics Science and Systems (RSS)}, June. 2016.

\bibitem[Weinberger and Saul(2009)]{weinberger2009distance}
K.~Q. Weinberger and L.~K. Saul.
\newblock Distance metric learning for large margin nearest neighbor
  classification.
\newblock \emph{Journal of Machine Learning Research}, 10\penalty0
  (Feb):\penalty0 207--244, 2009.

\bibitem[Xiang et~al.(2008)Xiang, Nie, and Zhang]{xiang2008learning}
S.~Xiang, F.~Nie, and C.~Zhang.
\newblock Learning a mahalanobis distance metric for data clustering and
  classification.
\newblock \emph{Pattern Recognition}, 41\penalty0 (12):\penalty0 3600--3612,
  2008.

\bibitem[Zampogiannis et~al.(2015)Zampogiannis, Yang, Ferm{\"u}ller, and
  Aloimonos]{zampogiannis2015learning}
K.~Zampogiannis, Y.~Yang, C.~Ferm{\"u}ller, and Y.~Aloimonos.
\newblock Learning the spatial semantics of manipulation actions through
  preposition grounding.
\newblock In \emph{2015 IEEE International Conference on Robotics and
  Automation (ICRA)}, pages 1389--1396. IEEE, 2015.

\end{thebibliography}
\addtolength{\textheight}{-12cm}   

\end{document}